# MULTI-TEMPORAL SPATIAL-SPECTRAL COMPARISON NETWORK FOR HYPERSPECTRAL ANOMALOUS CHANGE DETECTION

*Meiqi Hu[1], Chen Wu*[1], Bo Du[2]*

[1]State Key Laboratory of Information Engineering in Surveying, Mapping, and Remote Sensing, Wuhan University, Wuhan 430079, P. R. China
[2]School of Computer, Wuhan University, Wuhan 430072, P. R. China

## ABSTRACT

Hyperspectral anomalous change detection has been a challenging task for its emphasis on the dynamics of small and rare objects against the prevalent changes. In this paper, we have proposed a Multi-Temporal spatial-spectral Comparison Network for hyperspectral anomalous change detection (MTC-NET). The whole model is a deep siamese network, aiming at learning the prevalent spectral difference resulting from the complex imaging conditions from the hyperspectral images by contrastive learning. A three-dimensional spatial spectral attention module is designed to effectively extract the spatial semantic information and the key spectral differences. Then the gaps between the multi-temporal features are minimized, boosting the alignment of the semantic and spectral features and the suppression of the multi-temporal background spectral difference. The experiments on the "Viareggio 2013" datasets demonstrate the effectiveness of proposed MTC-NET.

## 1. INTRODUCTION

Hyperspectral image (HSI) has provided marvelous spectral details for precise object recognition and change analysis. And hyperspectral change detection has drawn more and more attention in recent years. Change detection refers to identify the dynamics or changes of phenomena at the same location from the remote sensing images acquired on different times [1], widely applied in land use and land change analysis, city expansion monitoring, etc. Hyperspectral anomalous change detection (HACD) mainly focuses on the appearance, disappearance, displacement, concealment of small and rare objects [2], which are mostly related with man-made motion. HACD aims at highlighting these easily overlooked changes for decision-makers, gaining extensive application in airborne defense, illegal construction, disaster emergency, etc.

HACD puts an emphasis on the changes of small objects. However, the complex imaging conditions cast fierce challenges for the change detection. The motion of sensors, the various atmosphere and illumination conditions have great impact on shaping the background spectrum, leading to prevalent changes. The violent spectral difference of the background induces abundant missing detection and false alarms.

The traditional HACD methods fall into two categories at different point of views. From the perspective of binary change detection, the anomalous changes arise from the dynamics of object of interest, e.g. mountain fires, military equipment, illegal construction. The representative method is characterized by projecting the multi-temporal HSIs into the same imaging conditions, aiming at minimizing the spectral differences of background and highlighting the anomalous changes in the residual images. The classic predictor-based approaches, taking the Chronochrome (CC) [3] for example, assumes the multi-temporal HSIs can be represented by a linear space-invariant observation model. And the spectral difference resulting from the altered imaging conditions could be modeled by the linear model and are suppressed on the residual map, thus the anomalous changes are highlighted. But the linear model gains less perfect performance under the complex and variant imaging conditions. And the ACDA [4] has offered an effective non-linear model based on auto-encoder is proposed to eliminate the spectral difference between multi-temporal HSIs. However, the ACDA takes only spectral information into consideration, leading to sparkle noise.

From the perspective of anomaly detection, the anomalous changes are regarded as the spectral anomalies generated by two single HSIs, where the displacement is thought as position altered anomaly, and the replacement as well as camouflage are looked as temporal anomalies. On the basis of the statistical distribution theory, most algorithms hold that the whole image could be considered as a certain distribution model. The Difference Reed-Xiaoli (Diff-RX) [5] detects those pixels that depart from the main distribution model from the difference images. And straight anomalous change detector [2] finds out the pixels varying from the assumed distribution based on the generalize likelihood ratio test. Nonetheless, the assumption can be inappropriate, producing inaccurate detection result.

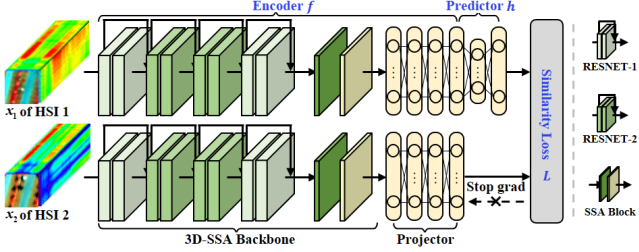

**Fig. 1.** The framework of propose MTC-NET.

It is thought that there exists low change semantic information for the prevalent change of multitemporal backgrounds for their similar semantic information. Whereas multitemporal anomalous changes areas hold high change semantic information due to different semantic information. The true anomalous changes are differentiated from the pseudo-changes from the point view of the semantic information comparison. Therefore, self-supervised learning (SSL) [6] provides a good potential in dealing with the HACD, since it is an unsupervised representation learning to discriminate the similarity of semantic information. Contrastive learning (CL) is a kind of discriminative SSL, aiming to attract the positive sample pairs and repulse the negative sample pairs. And the multi-temporal HSIs are naturally two different views of an image at the same location, regarded as the positive sample pairs. The SSL is employed to learn the spectral invariance from the spectral difference comparisons.

Inspired by this, we proposed a Multi-Temporal spatial-spectral Comparison Network for hyperspectral anomalous change detection (MTC-NET), which integrated with SimSiam [7] to learn the spatial and spectral feature representation from data comparison. A siamese deep neural network are designed to learn to discriminate the spectral difference of multitemporal backgrounds, where a three-dimensional (3D) spatial-spectral attention (3D-SSA) neural network is tailored to extract the invariant spectral features and align the projected features with similar semantic information. For those anomalous change areas with large change semantic information, the projected features are departed in the feature space. The proposed model is trained in an unsupervised way. For inference, the distance of the learned features is taken as the anomalous change intensity.

The paper is organized as follows. Section II will present the proposed method MTC-NET. The experiment results and analysis will be exhibited on Section III. And Section IV will conclude this paper.

## 2. METHODOLOGY

### 2.1 Framework

**Fig. 1** shows the framework of proposed MTC-NET. The whole network is a siamese deep neural network for both bi-temporal HSI inputs. The MTC-NET is composed of the encoder and the predictor. Moreover, encoder consists of 3D-SSA backbone and projector, where the 3D-SSA extracts the spatial semantic information and the spectral features from the hyperspectral input, and then the projector maps the extracted feature into the feature space. The predictor is used to promote the non-linear feature extraction ability of the whole model. The RESNET-1, RESNET-2 are different scaled three-dimensional residual blocks, designed to extract multi-scaled feature representations. And the spatial spectral attention block is employed to emphasize on the most violent spatial region and the most discriminated spectral wavelength. Since the two branches share the same weight, the spatial spectral attention block balance the different spatial and spectral focuses between the bi-temporal inputs. The similarity loss computes the extracted feature from one temporal input and another temporal input with stopping gradient, aligning the semantic information and the essence spectral features between bi-temporal backgrounds. After training, the feature learned by the 3D-SSA are used for anomalous change detection by a L2 distance measurement.

### 2.2 MTC-NET

The proposed MTC-NET is derived from the contrastive learning method SimSiam, which only demands positive samples for representation learning. And it is appropriate to assume most of the whole images are unchanged since there are rare and small changes happened on bi-temporal HSIs. And we adopt the positive sample strategy that the whole image is segmented into several non-overlapped patches. The bi-temporal HSI patches are mainly unchanged. However, the variant spectral features are influenced by the complex imaging conditions, leading to comprehensive data augmentation on bi-temporal HSIs.

Mathematically, $X_1 \in \mathbb{R}^{H \times W \times C}$ and $X_2 \in \mathbb{R}^{H \times W \times C}$ are defined as the bi-temporal HSIs. The two augmentation patches of the same location $x_1 \in \mathbb{R}^{m \times m \times C}$ and $x_2 \in \mathbb{R}^{m \times m \times C}$ are feed into the encoder $f$ to acquire projected features on the feature space. And one of the features is further processed by the predictor, where the output is matched with the other projected feature. Denoting that the two projected features as $z_1$ and $z_2$, where $z_1 = f(x_1)$ and $z_2 = f(x_2)$. And the output from the predictor is defined as $p_1$, where $p_1 = h(f(x_1))$. And the similarity loss $L$ computes the distance between the two output vectors $p_1$ and $z_2$. The similarity distance deploys the cosine similarity as $D$, which is defined as follows:

$$L = -\frac{1}{2}\big(D(p_1, \text{stopgrad}(z_2)) + D(p_2, \text{stopgrad}(z_1))\big) \quad (1)$$

$$D(p_1, \text{stopgrad}(z_2)) = \frac{p_1}{\|p_1\|_2} \cdot \frac{\text{stopgrad}(z_2)}{\|\text{stopgrad}(z_2)\|_2} \quad (2)$$

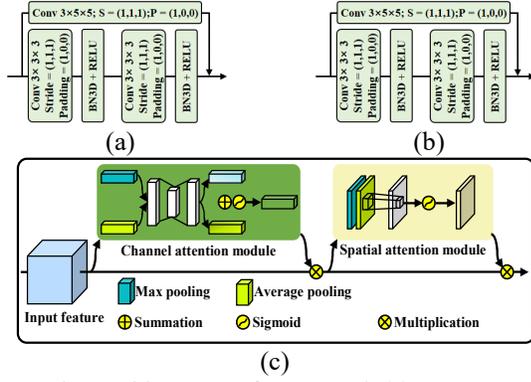

Fig. 2. The architecture of proposed (a) RESNET-1, (b) RESNET-2 and the (c) CBAM.

where $p_2 = h(f(x_2))$ are output vector of HSI 2. Noted that the stop gradient operation is the key point that drives the siamese network to learn the representation, where the parameter of the whole network and the representation are iteratively optimized to learn the features from the inputted positive sample pairs. And two similarity loss can be acquired with swapping the two inputted patches.

In details, the 3D-SSA are composed of multi-scaled three-dimensional residual blocks. As showed in **Fig. 2** (a), (b). The two residual blocks are composed of two 3D convolution layers, with 3D batchnorm and Relu activation function follows. Moreover, the RESNET-1 firstly possessed the input patches, where the 3D convolution slides from the spatial and spectral simultaneously and the strides are all equivalent to one, obtaining shallow features. And the first 3D convolutional layer in the RESNET-2 adopts stride equal to two on the spectral axis, aiming at condensing the spectral features and lessening the redundant spectral information. The down sample is employed for resizing the features as addition to the extracted features.

**Fig. 2** (c) represents the architecture of the spatial spectral attention block, employing the Convolutional Block Attention Module (CBAM) [8] to focus on the most violent spatial area and most discriminated spectral features. The projector is composed of three fully connected (FC) layers, with another two FC layers as the predictor, projecting the extracting the extracting feature on nonlinear feature space.

After the whole training, the feature extracted from the 3D-SSA are used for anomalous change detection, where L2 distance is adopted, as follows:

$$Loss\ map = \|3\text{D-SSA}(X_1) - 3\text{D-SSA}(X_2)\|_2 \qquad (3)$$

## 3. EXPERIMENTS

### 3.1 Data Description

In order to test the performance of proposed MTC-NET, two experiments are implemented on the classical HACD dataset "Viareggio 2013" [9]. Three HSIs are D1F12H1, D1F12H2 and D2F22H2, respectively. And the first experiment named "EX-1" is composed of D1F12H1 and D1F12H2, both of which are acquired on a similar imaging condition. And the second one named "EX-2", consists of D1F12H1 and D2F22H2, where there are big differences between the imaging conditions. The spectral solution is 1.2 nm, with 127 spectral bands, spatial size 450×375. The relative radiometric correction is operated on each experiment.

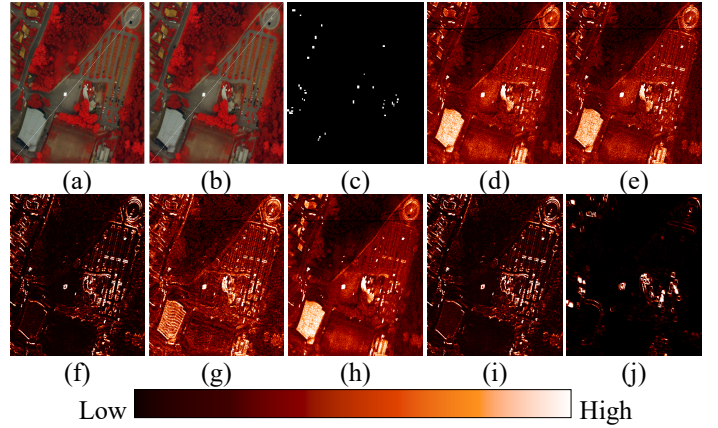

Fig. 3. The pseudo visualization of EX-1, (a) D1F12H1, (b) D1F12H2, (c) Reference change map. The anomalous change detection map of (d) CC, (e) CE, (f) USFA, (g) Diff_RX, (h) SACD, (i) SDHACD, (j) MTC-NET.

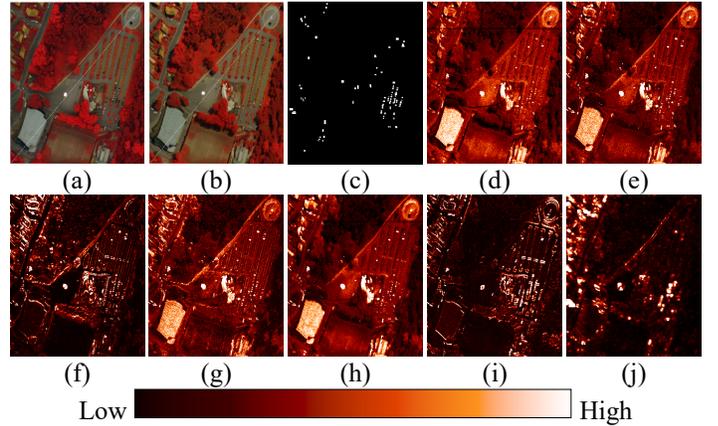

Fig. 4. The pseudo visualization of EX-2, (a) D1F12H1, (b) D1F12H2, (c) Reference change map. The anomalous change detection map of (d) CC, (e) CE, (f) USFA, (g) Diff_RX, (h) SACD, (i) SDHACD, (j) MTC-NET.

The experiment is implemented on pytorch. And the patch size $m$ adopted for training is 31×31. The training epoch is set as 100, with 128 as the batch size, and 0.05 as the initial learning rate with cosine decay. Another seven algorithms are deployed as comparison, including CC, Covariance Equalization (CE) [10], Unsupervised Slow Feature Analysis (USFA) [11], Diff-RX, Straight Anomalous Change Detector (SACD) and Simple Difference Hyperbolic Anomalous Change Detector (SDHACD).

### 3.2 Results and Analysis

The anomalous change detection result of EX-1 are showed in **Fig. 3**. Compared with the reference, the background area of the results of proposed MTC-NET are mostly in dark, indicating that the background is compressed. And most of anomalous change area are detected in high values. **Fig. 4** represents the result of EX-2. In the result of proposed MTC-NET, the background is mainly in dark, especially the bottom left square and the top right circle, indicating that the learned features of the bi-temporal backgrounds are highly aligned and the spectral differences are deeply compressed on the anomalous change detection map. For quantitative assessment, **Fig. 5** represents the ROC performance of two experiments. The closer the curve is to the top left, the better the performance the method obtains. It is observed that the AUC curve of proposed MTC-NET outperforms that of other methods for the EX-1. For EX-2, the MTC-NET in black acquires good effect under higher false alarms. And another evaluation index AUC is an index of comprehensive evaluation. The bigger it is, the better performance the method acquires. As TABLE I shows, the proposed MTC-NET obtains the highest AUC value on both experiments, demonstrating the effectiveness of proposed method.

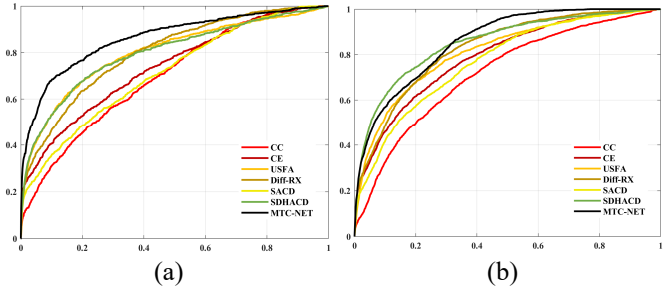

**Fig. 5.** ROC curves of (a) EX-1 and (b) EX-2.

**TABLE I THE AUC COMPARISON**

| Method | EX-1 | EX-2 |
|---|---|---|
| CC | 0.6991 | 0.7196 |
| CE | 0.7372 | 0.7896 |
| USFA | 0.8002 | 0.8127 |
| Diff_RX | 0.7993 | 0.8277 |
| SACD | 0.7128 | 0.7711 |
| SDHACD | 0.7978 | 0.8525 |
| MTC-NET | **0.8629** | **0.8594** |

## 4. CONCLUSION

In this paper, we have proposed an effective HACD method named MTC-NET which introduces the self-supervised learning to learn the prevalent spectral differences from the multi-temporal HSIs and align the learned features to compress the spectral difference between multi-temporal backgrounds. Two experiments are implemented on classical "Viareggio 2013" dataset, and the qualitive and quantitative results confirm the validity of proposed method.

## 5. ACKNOWLEDGMENT

This work was supported in part by National Natural Science Foundation of China under Grant T2122014 and 61971317, and in part by the Natural Science Foundation of Hubei Province under Grant 2020CFB594.